\documentclass[lettersize,journal]{IEEEtran}
\usepackage{amsmath,amsfonts}
\usepackage{algorithmic}
\usepackage{algorithm}
\usepackage{array}
\usepackage[caption=false,font=normalsize,labelfont=sf,textfont=sf]{subfig}
\usepackage{textcomp}
\usepackage{stfloats}
\usepackage{url}
\usepackage{verbatim}
\usepackage{graphicx}
\usepackage{cite}
\usepackage{xcolor}
\usepackage{multirow}
\usepackage{hyperref}
\begin{document}

\title{Local-Aware Global Attention Network for Person Re-Identification Based on Body and Hand Images}

%\author{IEEE Publication Technology,~\IEEEmembership{Staff,~IEEE,}
%        % <-this % stops a space
%\thanks{This paper was produced by the IEEE Publication Technology Group. They are in Piscataway, NJ.}% <-this % stops a space
%\thanks{Manuscript received April 19, 2021; revised August 16, 2021.}}

\author{Nathanael L. Baisa
%\author{Nathanael L. Baisa, Bryan Williams, Hossein Rahmani, Plamen Angelov, Sue Black  

\thanks{Nathanael L. Baisa is with the School of Computer Science and Informatics, De Montfort University, Leicester LE1 9BH, UK. Email: nathanael.baisa@dmu.ac.uk.}
%\thanks{Bryan Williams, Hossein Rahmani, Plamen Angelov and Sue Black are with the School of Computing and Communications, Lancaster University, Lancaster LA1  4WA, UK. Email: \{b.williams6, h.rahmani, p.angelov, sue.black\}@lancaster.ac.uk.}
}

%% The paper headers
%\markboth{Journal of \LaTeX\ Class Files,~Vol.~14, No.~8, August~2021}%
%{Shell \MakeLowercase{\textit{et al.}}: A Sample Article Using IEEEtran.cls for IEEE Journals}

%\IEEEpubid{0000--0000/00\$00.00~\copyright~2021 IEEE}
% Remember, if you use this you must call \IEEEpubidadjcol in the second
% column for its text to clear the IEEEpubid mark.

\maketitle

\begin{abstract}
Learning representative, robust and discriminative information from images is essential for effective person re-identification (Re-Id). In this paper, we propose a compound approach for end-to-end discriminative deep feature learning for person Re-Id based on both body and hand images. We carefully design the Local-Aware Global Attention Network (LAGA-Net), a multi-branch deep network architecture consisting of one branch for spatial attention, one branch for channel attention, one branch for global feature representations and another branch for local feature representations. The attention branches focus on the relevant features of the image while suppressing the irrelevant backgrounds. The global and local branches intends to capture global context and fine-grained information, respectively. A set of ablation study shows that each component contributes to the increased performance of the LAGA-Net. Extensive evaluations on four popular body-based person Re-Id benchmarks and two publicly available hand datasets demonstrate that our proposed method consistently outperforms existing state-of-the-art methods.

\end{abstract}


\section{Introduction} \label{sec:intro}

Person re-identfication (Re-Id), matching a particular person across different times, places, or cameras, has recently received a lot of attention from both industry and academia for different applications such as intelligent video surveillance. It is currently the main component of visual tracking in both single camera~\cite{Nat21} and multiple camera~\cite{RisTom18}. It is also similar to image retrieval in many aspects. Given a query image, person Re-Id ranks gallery images in terms of similarity to the query image. The image can be of a person's body, hand, face, etc. In this process, each image is represented with a feature embedding. Learning robust and discriminative feature representations is very crucial to overcome the many challenges the person Re-Id is facing. These challenges~\cite{WeiZhaGao18} include pose variations, occlusion, view point changes, lighting changes, background clutter, noisy labels, etc. Various efforts have been made to address these challenges~\cite{YeSheLin21}, for instance, considering whole body~\cite{WeiZhaXia14, ZheXiaZhi19, LuoGuLia20}, body parts~\cite{SuLiZha17, YifLiaYi18, WanYuaChe18} and attention mechanisms~\cite{BinWeiJia19, TiaShaJin19, ZhiCuiWen20} for learning robust and discriminative feature representations from person body images in uncontrolled environments for better matching.

Person Re-Id for biometric application has recently also received a lot of attention~\cite{JaiDebEng22}. Hand images, one of the primary biometric traits~\cite{DanEliRos16, BaiWilHosGPA21}, deliver discriminative features for biometric person recognition. Hand images not only have less variability when compared to other biometric modalities but also have strong and diverse features which remain relatively stable after adulthood~\cite{BaiWilHosGPA21, YimChaShu20, AttAkhCha21}. Because of this, there is a strong potential to investigate hand images captured by digital cameras for person recognition, especially for criminal investigation in uncontrolled environments since they are often the only available information in cases of serious crime such as sexual abuse.

In this work, we propose a compound approach for end-to-end discriminative deep feature representations learning for person Re-Id based on both body and hand images. We carefully design the Local-Aware Global Attention Network (LAGA-Net), a multi-branch deep network consisting of 4 branches to learn deep global, attentive and local (part-level) feature representations which are robust and discriminative enough for dealing with person Re-Id challenges. Each branch has its own merit within the goal of the proposed network architecture for learning robust and discriminative deep feature representations. Specifically, the attention branches, channel attention branch and spatial attention branch, focus on the relevant features of the image while suppressing the irrelevant backgrounds. The spatial attention branch incorporates relative positional encodings into spatial attention module to maintain translation equivariance. The global and local branches intends to capture global context and fine-grained information, respectively. By carefully designing this end-to-end compound method, we have shown that it is possible to effectively learn robust and discriminative feature representations for person Re-Id based on both body and hand images, unlike the previous methods~\cite{WeiZhaXia14, ZheXiaZhi19, LuoGuLia20,SuLiZha17, YifLiaYi18, WanYuaChe18,BinWeiJia19, TiaShaJin19, ZhiCuiWen20}. Our contributions can be summarized as follows.

%In this work, we propose a novel hand-based person recognition method, Multi-Branch with Attention Network (MBA-Net), by learning attentive deep feature representations from hand images captured by digital cameras for criminal investigations. Our contributions can be summarized as follows. %1) we propose a network by incorporating both channel and spatial attention modules in branches in addition to the global (without attention) branch for hand-based person recognition which is efficient computationaly and flexible in terms of the backbone architecture, 2) we include relative positional encodings into the spatial attention module, considering height and width independently, to capture the spatial positions of pixels in order to overcome the weakness of the attention mechanisms - equivariant to pixel shuffling, for efficiently recognizing suspects based on hand images, and 3) we make extensive evaluations on two large multi-ethnic and public hand datasets (11k hands~\cite{Mah19} and HD~\cite{KumXu16}  datasets). Our proposed method is trained in an end-to-end manner. 

\begin{enumerate}
\item We propose a multi-branch deep network by incorporating both channel and spatial attention modules in branches in addition to global (without attention) and local branches for person Re-Id based on both body and hand images which is efficient computationally and flexible in terms of the backbone architecture.
\item We include relative positional encodings into the spatial attention module, considering height and width independently, to capture the spatial positions of pixels in order to overcome the weakness of the attention mechanisms - equivariant to pixel shuffling, for efficiently re-identifying a person based on body and hand images.
%\item We incorporate the local (part-level) representation learning in addition to the global (with and without attention) features proposing a combined approach for robust discrimination of persons.
\item We make extensive evaluations on four popular body datasets: Market-1501~\cite{ZheSheTia15}, DukeMTMC-Re-ID~\cite{RisSolZou16}, CUHK03~\cite{WeiZhaXia14}, MSMT17~\cite{WeiZhaGao18} and two publicly available hand datasets: 11k~\cite{Mah19}, HD~\cite{KumXu16}, and LAGA-Net significantly outperforms existing state-of-the-art methods on these datasets.
\end{enumerate}

%We presented a preliminary idea of this work in~\cite{BaiWilHosMBA21}. In this work, we make more elaborate descriptions of our algorithm. Besides, we include local (part-level) information and hard mining triplet loss for learning more robust and discriminative deep feature representations. We also make extensive evaluations on the body person Re-Id benchmarks in addition to the hand datasets.

The rest of the paper is organized as follows. After the discussion of related work in Section~\ref{sec:RelatedWork}, the proposed method is described in Section~\ref{sec:proposedMethod} including the attention modules, the overall architecture of the LAGA-Net and the loss functions. The experimental results are analyzed and compared in Section~\ref{sec:experimentalResults} followed by the discussion in Section~\ref{sec:Discussion} and the main conclusion along with suggestion for future work in Section~\ref{sec:Conclusion}.

\section{Related Work}  \label{sec:RelatedWork}

\subsection{Body-based Person Re-Id}

Many person Re-Id methods have been proposed over the last few years, with more performance gain obtained by methods based on deep learning, using both supervised~\cite{SuLiZha17, YifLiaYi18, WanYuaChe18, BinWeiJia19, TiaShaJin19, ZhiCuiWen20} and unsupervised~\cite{XiaYixYu21, ShiShi21} learning approaches. Some of these works are based on learning global deep feature representations~\cite{WeiZhaXia14, ZheXiaZhi19, LuoGuLia20}. To overcome the performance limitations of the person Re-Id methods based on global features, researchers shifted their attention to learn deep local (part-level) feature representations by considering person poses~\cite{SuLiZha17}. In this case, external pose (skeleton) estimation methods have been used to leverage human part cues. However, this also comes with another disadvantage since the errors in the pose estimation can propagate to the re-identification stage. Mask is also used as external cues to remove the background clutters in pixel level to retrieve
body shape information~\cite{SonHuaOuy18}. Using these external methods bring additional computational burden. To overcome this, uniform partitioning of the images without relying on external methods was introduced for body-based person Re-Id in~\cite{YifLiaYi18, WanYuaChe18} and for hand-based person identification in~\cite{BaiWilHosGPA21}. Though this approach helped in gaining performance boost over the previous methods, the performance of the methods based on this approach is not sufficient to handle the challenges the person Re-Id methods are facing, probably due to pose misalignment problem. 

%Researchers have demonstrated that leveraging human part cues from external pose estimation to alleviate pose variations and learning feature representations can improve person re-identification based on images of the body~\cite{SuLiZha17}. However, the errors in the pose estimation can propagate to the identification which limits its performance. To allievate this problem, uniform partitioning for learning local (part-level) features has been proposed in~\cite{YifLiaYi18}, however, this method considered only the part-level features and in turn it has limited performance. 

Recently, self-attention mechanism, an integral component of Transformers~\cite{KhaNasHay21}, has received great attention in deep learning. The self-attention mechanism captures long-term information and interactions amongst all entities (e.g. pixels, channels, sequence elements, etc.) of the input data. It updates each pixel, for instance, by aggregating global information from all pixels in the input image. The attention mechanism has been used in~\cite{BinWeiJia19, TiaShaJin19, ZhiCuiWen20} for person Re-Id by considering both channel and spatial attentions which compute correlations between all the channels and all the pixels of the input feature map, respectively. However, these methods used the attention modules across the entire network layers which make it computationally inefficient since the self-attention computation is very expensive if the dimension of the input data is very large. Furthermore, these methods have limited performance when applied to person body and hand images. By default, the attention mechanism does not model relative or absolute position information. To overcome this, a relative position representation of a sequence element with respect to its neighbours has been proposed in~\cite{ShaJakAsh18} for natural language processing, and later on incorporated into a standalone global attention-based deep network for images without using convolutions for modeling pixel interactions~\cite{SheBelVem20}. Unlike these methods, we use the attention mechanism along with the convolution operations by applying the attention modules only to low-resolution feature maps in later stages of a deep network in branches along with global (without attention) and local branches for better computational efficiency as well as accuracy for person Re-Id based on not only body images but also hand images.

\subsection{Hand-based Person Re-Id}
Both traditional and deep learning approaches have been combined in~\cite{Mah19,YimChaShu20} to develop person identification using hand images. After training a convolutional neural network (CNN) on digital hand images (RGB), the method in~\cite{Mah19} used the network as a feature extractor to obtain CNN-features which have been fed into a set of support vector machine (SVM) classifiers. The work in~\cite{YimChaShu20} used rather similar approach with additional data type for fusion, near-infrared (NIR) images. These methods are not an end-to-end. An end-to-end approach considering both horizontal and vertical uniform partitioning has been proposed in~\cite{BaiWilHosGPA21}. However, all these methods have a limited performance. Unlike these methods, our proposed method is a compound approach considering multi-branch network architecture which is efficient not only on body images but also on hand images for re-identifying individuals.

\section{Proposed Method} \label{sec:proposedMethod}

In this section, we introduce the two attention modules followed by the overall architecture of the LAGA-Net and the used loss functions. The goal of the attention modules is to supress irrelevant backgrounds while focusing on discriminative information of person appearances.

\subsection{Channel Attention Module}

Channel attention module (CAM) aims to aggregate channel-wise feature-level information since some channels in higher convolutional layers are semantically related i.e. CAM computes the correlations between all the channels. The structure of CAM is given in Fig.~\ref{fig:CAM}. Given the input feature map $\mathbf{E}_i \in \mathbb{R}^{C \times H \times W}$ where C, H, W are the number of channels, height and width of the feature map, respectively, we first reshape it to produce the matrices of keys, queries and values, respectively, as $\mathbf{K} \in \mathbb{R}^{C \times HW}$, $\mathbf{Q} \in \mathbb{R}^{C \times HW}$ and $\mathbf{V} \in \mathbb{R}^{C \times HW}$. Then, the global channel attention map $\mathbf{A}_c \in \mathbb{R}^{C \times C}$ is computed using the dot-product of the query with all keys as

\begin{equation}
    \mathbf{A}_c = \rho (\mathbf{K} \mathbf{Q}^T)
\label{eq:Ac}
\end{equation}
\noindent where $\mathbf{Q}^T$ denotes the matrix transpose of $\mathbf{Q}$, and $\rho$ represents the softmax normalization along each row separately. The self-attended output feature map $\mathbf{E}_o \in \mathbb{R}^{C \times H \times W}$ for the channel attention is given by

\begin{equation}
    \mathbf{E}_o = \gamma (\mathbf{A}_c \mathbf{V}) + \mathbf{E}_i
\label{eq:Ac}
\end{equation}
\noindent where $\gamma$ is initialized as 0 and gradually learns to assign more weight to adjust the impact of the CAM. $\mathbf{A}_c \mathbf{V}$ means that the matrix of values $ \mathbf{V}$ is weighted by the attention score $\mathbf{A}_c$.

\subsection{Spatial Attention Module with Relative Positional Encodings}

\begin{figure*}[htbp]%[!htb] %[t]%[!h]
  \begin{center}
   \subfloat[] %[Channel Attention Module (CAM)]
  {\label{fig:CAM} \includegraphics[width=0.80\linewidth]{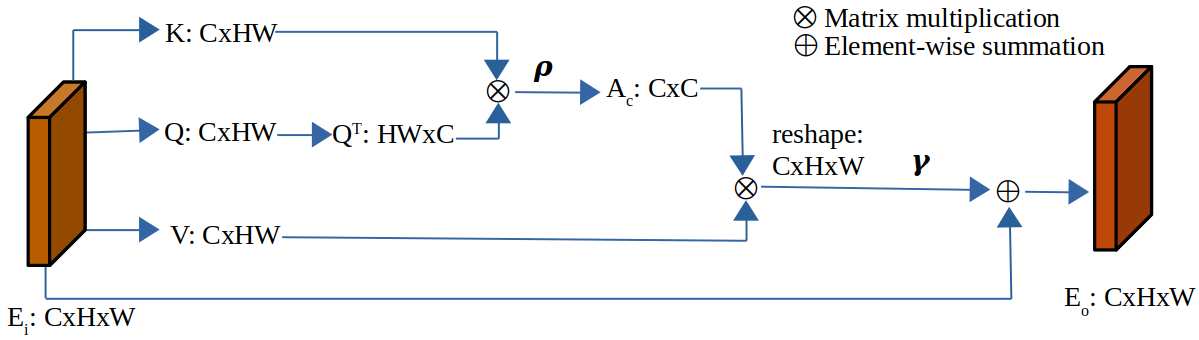}} \\%& %\\ %height=0.46
  \subfloat[] %[Spatial Attention Module with Relative Positional Encodings (SAM-RPE)]
  {\label{fig:SAM-RPE} \includegraphics[width=0.80\linewidth]{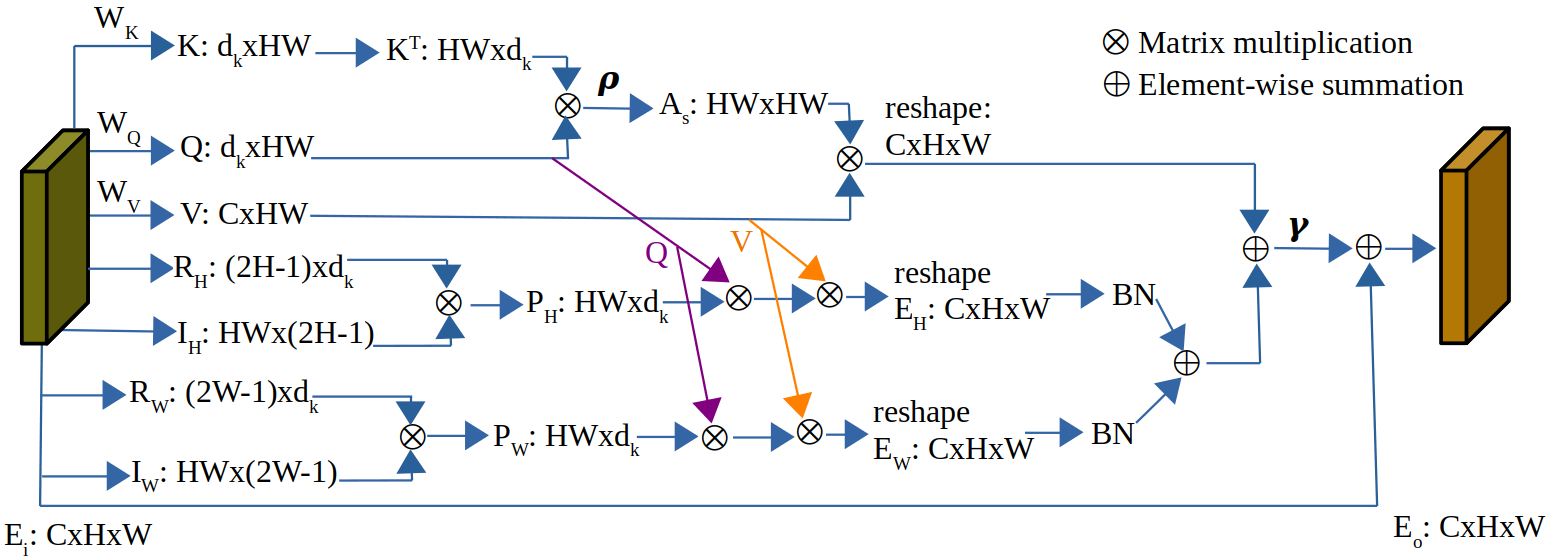}}  %& %\\ %height=0.18
  \end{center}
   \caption{Attention modules used in our proposed LAGA-Net: (a) Channel Attention Module (CAM), (b) Spatial Attention Module with Relative Positional Encodings (SAM-RPE).}
  \label{fig:CAM_SAM-RPE}
\end{figure*}

The goal of spatial attention module is to aggregate the semantically similar pixels in the spatial domain of the input feature map. Though the spatial attention mechanism attends to the entire input feature map based on content (pixel values), it does not take into account the spatial positions of pixels which makes it equivariant to pixel shuffling. To overcome this, we incorporate the relative positional encodings along the rows (height) and columns (width), which is computationally efficient, so that it maintains translation equivariance i.e. translating (shifting) the input pixel also translates the output pixel by the same amount. The structure of the Spatial Attention Module with Relative Positional Encodings (SAM-RPE) is given in Fig.~\ref{fig:SAM-RPE}. 
Given the input feature map $\mathbf{E}_i \in \mathbb{R}^{C \times H \times W}$, the matrices of keys $\mathbf{K} \in \mathbb{R}^{d_k \times HW}$, queries $\mathbf{Q} \in \mathbb{R}^{d_k \times HW}$ and values $\mathbf{V} \in \mathbb{R}^{C \times HW}$ are obtained by transforming it through defined learnable weight matrices $\mathbf{W}_K$, $\mathbf{W}_Q$ and $\mathbf{W}_V$, respectively, where $d_k = \frac{C}{8}$ is the channels dimension of the keys and queries. The learnable weight matrices $\mathbf{W}_K$, $\mathbf{W}_Q$ and $\mathbf{W}_V$ are implemented using independent pointwise ($1 \times 1$) convolution layers with batch normalization and ReLU activation. Thus, the global spatial attention map $\mathbf{A}_s \in \mathbb{R}^{HW \times HW}$ is computed as

\begin{equation}
    \mathbf{A}_s = \rho (\mathbf{K}^T \mathbf{Q})
\label{eq:Ac}
\end{equation}
\noindent where $\mathbf{K}^T$ denotes the matrix transpose of $\mathbf{K}$, and $\rho$ represents the softmax normalization along each row separately.

We consider height (row) and width (column) attentions due to the relative spatial positions for computational efficiency. To compute these relative positional attentions, we first need to represent relative shifts along the height or width of the input feature map. Let a relative position embedding for the height that needs to be learned be $\mathbf{R}_H \in \mathbb{R}^{(2H-1) \times d_k}$ where $H$ is the height of the input feature map and $d_k =\frac{C}{8}$ is the number of channels. A possible vertical shift, from $-(H-1)$ to $H-1$, corresponds to each row of $\mathbf{R}_H$. The relative shifts in the matrix $\mathbf{R}_H$ need to be represented using absolute shifts. To do this, the re-indexing tensor $\mathbf{I}^H \in \mathbb{R}^{H \times W \times (2H-1)}$, which is used as a mask, can be defined as

\begin{equation}
    \mathbf{I}^H_{h,i,r} =
\begin{cases}
    1,  &  \text{if}~ i-h = r~~ \& ~~ |i-h |  \leq H \\
    0,  & \text{otherwise}
\end{cases}
\label{eq:labelSmoothing}
\end{equation}
\noindent where $h \in \{0, ..., H-1\}$, $i \in \{0, ..., W-1\}$ and $r \in \{-(H-1), ..., 0, ..., H-1\}$.

By reshaping $\mathbf{I}^H$ to $\mathbf{I}_H \in \mathbb{R}^{HW \times (2H-1)}$, a position embedding tensor with indices of absolute shifts  for the height $\mathbf{P}_H \in \mathbb{R}^{HW \times d_k}$ is given by 

\begin{equation}
    \mathbf{P}_H = \mathbf{I}_H \mathbf{R}_H
\label{eq:Ph}
\end{equation}
\noindent Then, the self-attended feature map $\mathbf{E}_H \in \mathbb{R}^{C \times H \times W}$ corresponding to the height relative position embedding $\mathbf{R}_H$, which is used as keys implicity ($\mathbf{P}_H$ explicitly), is computed as 

\begin{equation}
    \mathbf{E}_H = \mathbf{V} (\mathbf{P}_H \mathbf{Q})
\label{eq:Ph}
\end{equation}
\noindent where $\mathbf{P}_H \mathbf{Q}$ corresponds to the height relative positional attention.

The relative position embedding for the width $\mathbf{R}_W \in \mathbb{R}^{(2W-1) \times d_k}$, the re-indexing tensor $\mathbf{I}_W \in \mathbb{R}^{HW \times (2W-1)}$ and its corresponding self-attended feature map $\mathbf{E}_W \in \mathbb{R}^{C \times H \times W}$  can be obtained with similar approach to the above height formulation since they are symmetric.

Thus, the final self-attended output feature map $\mathbf{E}_o \in \mathbb{R}^{C \times H \times W}$ for the SAM-RPE is given by

\begin{equation}
    \mathbf{E}_o = \gamma (\mathbf{V} \mathbf{A}_s + BN(\mathbf{E}_H) + BN(\mathbf{E}_W)) + \mathbf{E}_i
\label{eq:Ac}
\end{equation}
\noindent where $\gamma$ is initialized as 0 and gradually learns to assign more weight to adjust the impact of the SAM-RPE, and $BN$ is a batch normalization.

%\begin{figure}
%  \centering
%  \begin{tabular}[b]{c}
%    \includegraphics[width=.8\linewidth]{images/CAM.png} \\
%    \small (a)
%  \end{tabular} \qquad  \\
%  \begin{tabular}[b]{c}
%    \includegraphics[width=.8\linewidth]{images/SAM.png} \\
%    \small (b)
%  \end{tabular}
%  \caption{Attention modules (a)~Channel Attention Module (CAM) (b)~Spatial Attention Module with Relative Positional Encodings (SAM-RPE).}
%  \label{fig:CAM_SAM-RPE}
%\end{figure}
%\noindent

%\begin{figure*}[htbp]%[!htb] %[t]%[!h]
%  \begin{center}
%   \subfloat[Channel Attention Module (CAM)]
%  {\label{fig:CAM} \includegraphics[width=0.80\linewidth]{images/CAM_laga.png}} \\%& %\\ %height=0.46
%  \subfloat[Spatial Attention Module with Relative Positional Encodings (SAM-RPE)]
%  {\label{fig:SAM-RPE} \includegraphics[width=0.80\linewidth]{images/SAM_laga.png}}  %& %\\ %height=0.18
%  \end{center}
%   \caption{Attention modules, CAM and SAM-RPE, used in our proposed LAGA-Net.}
%  \label{fig:CAM_SAM-RPE}
%\end{figure*}
%\noindent

\noindent

\subsection{Network Architecture Overview}

\begin{figure*}[t]%[htbp] %[t]%[!htb] %[t]%[!h]
\begin{center}
  \includegraphics[width=0.8\linewidth]{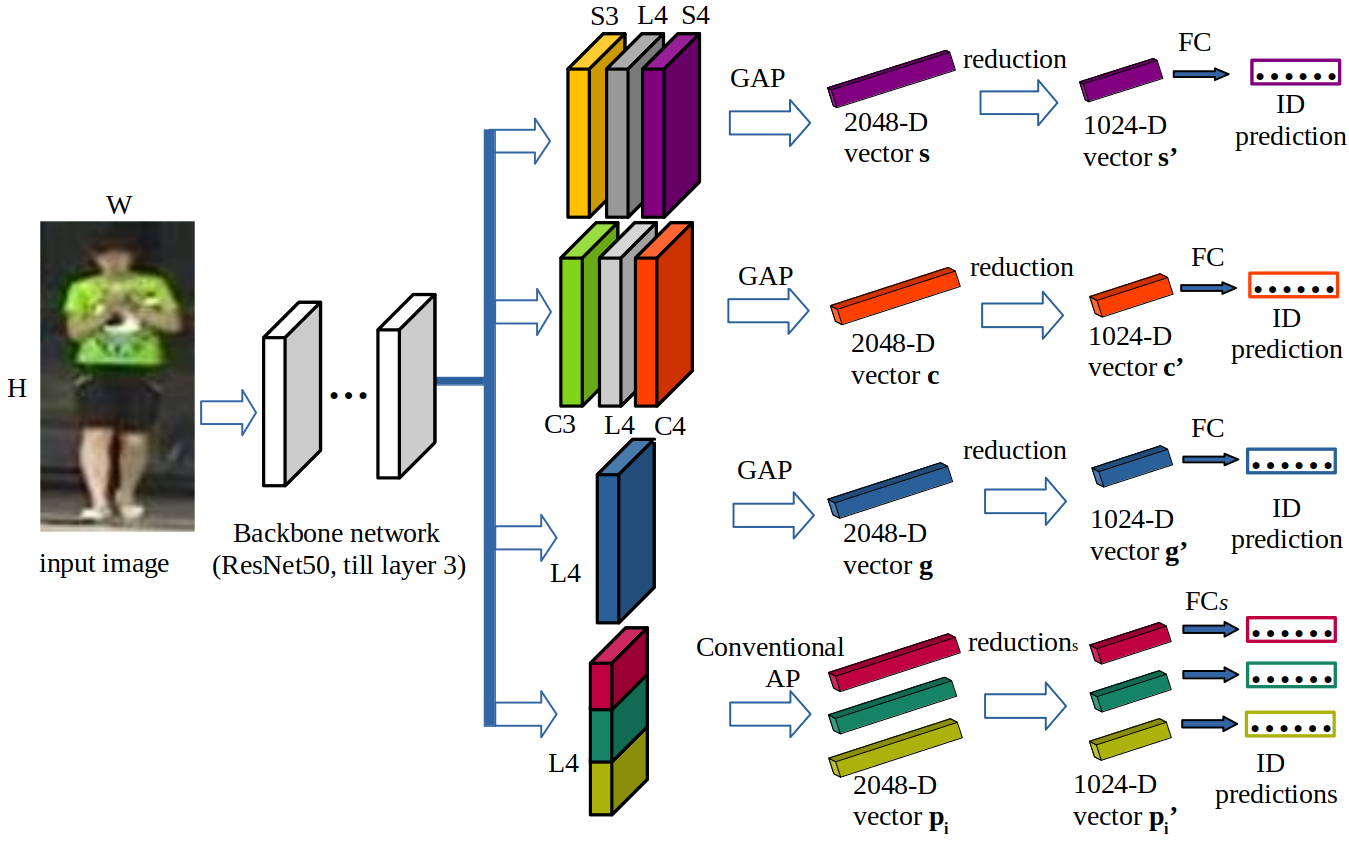} \\
\end{center}
   \caption{Structure of LAGA-Net. Four separate 3D tensors (one for spatial attention branch, one for channel attention branch, one for global branch and the other for local branch) are obtained by passing the input image through the stacked convolutional layers from the backbone network. S3 and S4 are the SAM-RPE after layer 3 and layer 4 (L4) of the ResNet50, respectively. Similarly, C3 and C4 are the CAM after layer 3 and layer 4 of the ResNet50, respectively. Three horizontal partitions (stripes) are also performed on L4 to produce the local branch. Given an input image, six separate 2048-D column feature vectors are obtained by passing it through the backbone network with the 4 branches (the local branch has 3 horizontal stripes). Each classifier predicts the identity (ID) of the input image during training. In case of hand-based person Re-Id, hand input images are used.}
\label{fig:LAGAdiagram}
\end{figure*}

The overall architecture of the proposed LAGA-Net is given in Fig.~\ref{fig:LAGAdiagram}. It incorporates two complementary attention modules, channel and spatial. These attention modules are used at higher level of the network, for computational efficiency, in branches along with the global (without attention) branch and the local branch which is obtained by performing uniform horizontal partitioning. As a backbone network, we use ResNet50~\cite{HeZhaSun16} pretrained on ImageNet due to its precise architecture with competitive performances in some person Re-Id works~\cite{YifLiaYi18, WanYuaChe18, TiaShaJin19, ZhiCuiWen20}. Obviously, any network designed for image classification can be adapted, for example Inception network~\cite{ChrSerVin17} and DenseNet~\cite{HuaLiuVan17}. We keep the structure of the original ResNet50 before layer 3 (inclusive) remain the same when we modify the backbone network to produce the LAGA-Net. We create 4 independent branches just after the layer 3 of the ResNet50 to incorporate the channel and the spatial (with relative positional encodings) attention modules in branches by keeping one (without attention) global branch and one additional local branch for which we generate 3 horizontal stripes uniformly from the output feature map.  

%For the channel, spatial and without attention global branches, a Global Average Pooling (GAP) layer is used to summarize the 3D tensor of activations to form a 2048-dimensional column feature vectors $\textbf{c}$, $\textbf{s}$ and $\textbf{g}$, respectively.

%For the local branches, we change from the GAP layer to conventional average pooling (AP) layer to create uniform horizontal partitions on the 3D tensor of activations to learn part-level features $\textbf{p}_i$ where $l \in [1, 3]$; the total number of partitions used is 3.

\noindent \textbf{Spatial attention branch:} This branch aggregates the semantically similar pixels in the spatial domain of the input feature map and it uses a Global Average Pooling (GAP) layer to summarize the 3D tensor of activations to form a 2048-dimensional column feature vector $\textbf{s}$.

\noindent \textbf{Channel attention branch:} This branch aggregates the correlations between all the channels of the input feature map and it uses the GAP layer to summarize the 3D tensor of activations to form a 2048-dimensional column feature vector $\textbf{c}$.

\noindent \textbf{Global branch:} This branch aims to maintain global context information for discriminative feature learning, and the GAP layer is used to summarize the 3D tensor of activations to form a 2048-dimensional column feature vector $\textbf{g}$.

\noindent \textbf{Local branch:} For this branch, we change from the GAP layer to conventional average pooling (AP) layer to create uniform horizontal partitions (stripes) on the 3D tensor of activations to learn 2048-dimensional part-level features $\textbf{p}_i$ where integer $i \in [1, 3]$; the total number of partitions used is 3.

%For each branch (for each stripe in the case of the local branch), a new fully-connected layer (FC), batch normalization (BN), leaky rectified lineat unit (LReLU) and dropout with probability of 0.5 to reduce possible over-fitting are employed to process 2048-dimensional column feature vectors obtained after the GAP and conventional AP layers which in turn are fed into the classification layers. Each classification layer, which is implemented using a FC layer followed by a softmax function, predicts the identity (ID) of each input. In addition, we change the last stride from 2 to 1 in the backbone network i.e. remove the last spatial down-sampling operation, which increases the size of the tensor of each branch for improved performance as observed in~\cite{LuoGuLia20}.

Each reduction layer for each branch (for each stripe in the case of the local branch) is implemented using a new fully-connected layer (FC), batch normalization (BN), leaky rectified linear unit (LReLU) and dropout with probability of 0.5 to reduce possible over-fitting.  The reduction layers are employed to convert 2048-dimensional column feature vectors obtained after the GAP and conventional AP layers to 1024-dimensional feature vectors which in turn are fed into the classification layers. Each classification layer, which is implemented using a FC layer followed by a softmax function, predicts the identity (ID) of each input. In addition, we change the last stride from 2 to 1 in the backbone network i.e. remove the last spatial down-sampling operation, which increases the size of the tensor of each branch for improved performance as observed in~\cite{LuoGuLia20}.

\noindent

\subsection{Loss Functions}

The LAGA-Net is optimized during training by minimizing the loss function $\mathcal{L}$ consisting of the sum of cross-entropy losses over the 6 ID predictions for identification (classification) and the sum of hard mining triplet losses over the 6 ID predictions for metric learning i.e. each classifier predicts the identity of the input image as shown in Fig.~\ref{fig:LAGAdiagram}. The total loss function $\mathcal{L}$ is given by

\begin{equation}
    \mathcal{L} = \sum_{l=1}^6 \mathcal{L}_{l,xent} + \beta \sum_{l=1}^6 \mathcal{L}_{l, triplet} 
\label{eqn:totalLoss}
\end{equation}
\noindent where $\beta$ is a hyperparameter balancing the two types of losses. %and is set to 0.1. %The margin parameter for triplet loss is set to 1.2 ($\alpha = 1.2$).

For the learned features $\textbf{f}_i$ (for sample $i$), the cross-entropy loss (softmax loss) with label smoothing~\cite{SzeVanIof16} is given as: 

%\begin{equation}
%    \mathcal{L}_{l,xent} = -\sum_{i=1}^N q_{y_i} \log \frac{e^{\textbf{W}^T_{y_i} \textbf{f}_i + b_{y_i}}}{\sum_{c=1}^C e^{\textbf{W}^T_c \textbf{f}_i + b_c}} 
%\label{eqn:xent}
%\end{equation}

\begin{equation}
    \mathcal{L}_{l,xent} = -\frac{1}{N}\sum_{i=1}^N \sum_{j=1}^C q_{y_{i,j}} \log \frac{e^{\textbf{W}^T_{j} \textbf{f}_i + b_{j}}}{\sum_{c=1}^C e^{\textbf{W}^T_c \textbf{f}_i + b_c}} 
\label{eqn:xent}
\end{equation}
\noindent where N is the batch-size, C is the number of classes (identities) in the training dataset, and $\textbf{W}_c$ and $b_c$ are weight vector and bias for class $c$, respectively. Note that $z_c = \textbf{W}^T_c \textbf{f}_i + b_c$ are the logits or unnormalized probabilities. The ground-truth distribution over labels $q_{y_{i,j}}$ by including label smoothing can be given as

\begin{equation}
    q_{y_{i,j}} =
\begin{cases}
    1 - \frac{C-1}{C} \epsilon, & \text{if } y_{i,j} = y\\
    \frac{1}{C} \epsilon,              & \text{otherwise}
\end{cases}
\label{eq:labelSmoothing}
\end{equation}
\noindent where $y$ is ground-truth label (for sample $i$ and class $j$) and $\epsilon$ is a smoothing value. 

Similarly, the batch hard mining triplet loss~\cite{AlexLucBas17} is given as follows:
\begin{equation}
\begin{array} {lll}
    \mathcal{L}_{l,triplet} &= \sum_{i=1}^P \sum_{a=1}^K \bigg[\alpha ~+~ \overbrace{\max_{p=1 \dots K} \|\textbf{f}_a^{(i)} - \textbf{f}_p^{(i)}\|_2 }^{hardest~positive}  \\& -  \underbrace{\min_{n=1 \dots K, j=1 \dots P, j \neq i} \|\textbf{f}_a^{(i)} - \textbf{f}_n^{(j)}\|_2}_{hardest~negative} \bigg]_+
\end{array}
\label{eqn:triplet}
\end{equation}
\noindent where $[A]_+ = \max(A, 0)$, $\alpha$ is the margin hyperparameter that controls the distance differences of intra and inter classes, and $\textbf{f}_a^{(i)}$, $\textbf{f}_p^{(i)}$, $\textbf{f}_n^{(i)}$ are the features extracted from anchor, positive and negative samples, respectively. The positive and negative samples refer to the persons with same or different identity with the anchor. The candidate triplets are constructed by the furthest positive and closest negative sampled pairs. These are basically the hardest positive and hardest negative pairs in a mini-batch $N = PK$ with $P$ selected identities and $K$ instances (images) per identity. In our experimental settings, we use $\beta = 0.1$, $\alpha = 1.2$, $P = 5$, $K = 4$ and $N = PK = 20$.

For both losses, we use the learned embeddings $\textbf{\'s}$, $\textbf{\'c}$, $\textbf{\'g}$, $\textbf{\'p}_1$, $\textbf{\'p}_2$ and $\textbf{\'p}_3$ (in place of $\textbf{f}$ in the Eqns.~(\ref{eqn:xent}) and (\ref{eqn:triplet})) during training (see in Fig.~\ref{fig:LAGAdiagram}). However, during testing, we concatenate all the 2048-D feature vectors of the 6 branches as the final feature embedding, just after the GAP and conventional AP i.e. $\mathcal{F} = [\textbf{s}, \textbf{c}, \textbf{g}, \textbf{p}_1, \textbf{p}_2, \textbf{p}_3]$ which becomes 12288-D feature vector, and then compare feature vector of each query image with gallery feature vectors using cosine distance. In fact, we put image $\textbf{I}$ and its horizontally flipped image $\textbf{I}'$ into the model, and get their embeddings $\mathcal{F}$ and $\mathcal{F}'$. Then their mean feature $\frac{\mathcal{F} + \mathcal{F}'}{2}$ is used as the embedding of image $\textbf{I}$ during testing which improves the Re-Id performance.

\section{Experiments}  \label{sec:experimentalResults}

\subsection{Datasets}

In this section, we describe both body person Re-Id datasets such as Market-1501~\cite{ZheSheTia15}, DukeMTMC-Re-ID~\cite{RisSolZou16}, CUHK03~\cite{WeiZhaXia14} and MSMT17~\cite{WeiZhaGao18}, and hand datasets such as 11k hands dataset~\cite{Mah19} and Hong Kong Polytechnic University Hand Dorsal (HD) dataset~\cite{KumXu16}.

\textbf{Market-1501}~\cite{ZheSheTia15} is captured by 6 cameras comprising 32,668 labeled images of 1,501 identities. 751 identities (12,936 images) are used for training whereas the rest are used for testing. The testing data is sub-divided into test probe (query) set and test gallery set. The test probe set has 3,368 images of 750 identities, while 2,793 additional distractors are also included into the test gallery set. The overall statistics of body-based person Re-Id datasets used in this paper is given in Table~\ref{tbl:BodyDatasets}.

\textbf{DukeMTMC-Re-ID}~\cite{RisSolZou16} is captured by 8 cameras containing 36,411 images of 1,812 identities. Among these identities, 1,404 identities appear in more than 2 cameras while 408 identities (distactors) appear in only one camera. 702 identitis are randomly chosen for training from the 1,404 identities and the rest are used for testing. One query image for each identity per camera is chosen from the testing set for the probe set while all remaining images including the distractors are used for gallery set.

\textbf{CUHK03}~\cite{WeiZhaXia14} contains 13,164 images of 1,467 persons, and each identity only appears in two disjoint camera views. The new training and testing protocol proposed in~\cite{ZhuLiaDon17} is adopted, in which 767 identities are used for training and 700 for testing. Both labeled and detected bounding boxes are given in CUHK03, we perform experiments on the labeled (L) bounding boxes.

\textbf{MSMT17}~\cite{WeiZhaGao18} is captured by a 15-camera network (12 outdoor, 3 indoor) and is currently the largest publicly available person Re-Id dataset with 126,441 images of 4,101 identities. The training set contains 32,621 images with 1041 persons (identities), while the test set contains 93,820 images with 3060 persons. For the test set, 11,659 images are randomly selected as query, and the other 82161 images are used as gallery i.e. we use the training-testing split of~\cite{WeiZhaGao18}. The video is collected with different weather conditions at three-time slots (morning, noon, afternoon). The annotations include camera IDs, weathers and time slots. Thus, the MSMT17 is significantly more challenging than the other three due to its massive scale, more complex and dynamic scenes.

\textbf{11k hands\footnote{\url{https://sites.google.com/view/11khands}} dataset}~\cite{Mah19} has 190 subjects (identities). We use the same partitioning strategy of the dataset as in~\cite{BaiWilHosGPA21}. As in~\cite{BaiWilHosGPA21}, this dataset is divided into right dorsal, left dorsal, right palmar and left palmar sub-datasets to train a hand-based person Re-Id (recognition) model. After excluding accessories and dividing the dataset as in~\cite{BaiWilHosGPA21},  right dorsal has 143 identities, left dorsal has 146, right palmar has 143 and left palmar has 151 identities. The first half and the second half of each sub-dataset based on identity are used for training and testing, respectively. For instance, for right dorsal, the first 72 identities are used for training and the last 71 identities are used for testing. Similarly, the first 73, 72 and 76 identities are used for training phase for left dorsal, right palmar and left palmar sub-datasets, respectively. The remaining identities of each sub-dataset (73 for left dorsal, 71 for right palmar, 75 for left palmar) are used for testing. From each identity of the test set of each sub-dataset, we randomly choose one image and put in a common gallery for all the 11k sub-datasets. The remaining images of each identity of the test set of each sub-dataset are used as a query set for that sub-dataset. Accordingly, the 11k gallery has 290 images and the query has 971 images for right dorsal, 988 images for left dorsal, 917 images for right palmar and 948 images for left palmar. A randomly chosen
image of each identity of the training set of each sub-dataset is used as a validation for monitoring the training process. This procedure is repeated for 10 times and the average performance is reported. The overall statistics of hand-based person Re-Id datasets used in this paper is given in Table~\ref{tbl:HandDatasets}.

\textbf{HD\footnote{\url{http://www4.comp.polyu.edu.hk/~csajaykr/knuckleV2.htm}} dataset}~\cite{KumXu16} has 502 identities. We use the same partitioning strategy of the dataset as in~\cite{BaiWilHosGPA21}. The first 251 identities and the second 251 identities of the dataset are used for training and testing, respectively. From each identity of the test set of this dataset, one image is randomly chosen and put in a gallery and the rest are used as a query (probe). Unlike the 11k dataset, the HD dataset has additional images of 213 subjects, which lack clarity or do not have second minor knuckle patterns, and are added to the HD gallery. Accordingly, the gallery for the HD dataset has 1593 images and the query has 1992 images. A randomly chosen image of each identity of the training set of the dataset is used as a validation for monitoring the training process. This procedure is repeated for 10 times and the average performance is reported.

\begin{table*} [htbp]%[!htb]%[!h]%[tb]
\caption{\normalfont{Statistics of body-based person Re-Id datasets used in this paper: Market-1501~\cite{ZheSheTia15}, DukeMTMC-Re-ID~\cite{RisSolZou16}, CUHK03~\cite{WeiZhaXia14} and MSMT17~\cite{WeiZhaGao18}. Number of identities (ids), number of images and number of cameras are shown for train set, query set and gallery set of each dataset.}}
\label{tbl:BodyDatasets}
\begin{center}
  \begin{tabular}{|l|lll|lll|lll|lll|}
    \hline
    \multirow{2}{*}{Subset} & 
      \multicolumn{3}{c|}{Market-1501} & 
      \multicolumn{3}{c|}{DukeMTMC}    &
      \multicolumn{3}{c|}{CUHK03 (L)}  &
      \multicolumn{3}{c|}{MSMT17}     \\
      \cline{2-13}
    & \# ids & \# images & \# cameras & \# ids & \# images & \# cameras & \# ids & \# images & \# cameras & \# ids & \# images & \# cameras \\
    \hline
    Train & 751 & 12,936 & 6 & 702 & 16,522 & 8 & 767 & 7,365 & 2 & 1,041 & 30,248 & 15 \\
    Query & 750 & 3,368 & 6 & 702 & 2,228 & 8 & 700 & 1,400 & 2 & 3,060 & 11,659 & 15 \\
    Gallery & 751 & 15,913 & 6 & 1,110 & 17,661 & 8 & 700 & 5,332 & 2 & 3,060 & 82,161 & 15 \\
	\hline 
  \end{tabular}
\end{center}
%\caption{Statistics of body-based person Re-Id datasets used in this paper: Market-1501~\cite{ZheSheTia15}, DukeMTMC-Re-ID~\cite{RisSolZou16}, CUHK03~\cite{WeiZhaXia14} and MSMT17~\cite{WeiZhaGao18}. Number of identities (ids), number of images and number of cameras are shown for train set, query set and gallery set of each dataset.}
%\label{tbl:BodyDatasets}
\end{table*}
\noindent

\begin{table*} [htbp]%[!htb]%[!h]%[tb]
\caption{\normalfont{Statistics of hand-based person Re-Id datasets used in this paper: 11k~\cite{Mah19} and HD~\cite{KumXu16}. Number of identities (ids) and number of images are shown for train set, query set and gallery set of each dataset. Only one camera is used to capture each of these datasets.}}
\label{tbl:HandDatasets}
\begin{center}
  \begin{tabular}{|l|ll|ll|ll|ll|ll|}
    \hline
    \multirow{2}{*}{Subset} & 
      \multicolumn{2}{c|}{D-r of 11k} & 
      \multicolumn{2}{c|}{D-l of 11k} & 
      \multicolumn{2}{c|}{P-r of 11k} &
      \multicolumn{2}{c|}{P-l of 11k} &
      \multicolumn{2}{c|}{HD} \\
      \cline{2-11}
    & \# ids & \# images & \# ids & \# images & \# ids & \# images & \# ids & \# images & \# ids & \# images \\
    \hline
	Train & 72 & 962 & 73 & 808 & 72 & 977 & 76 & 1,004 & 251 & 2,407 \\ 
	Query & 71 & 971 & 73 & 988 & 71 & 917 & 75 & 948 & 251 & 1,992 \\
	Gallery & 290 & 290 & 290 & 290 & 290 & 290 & 290 & 290 & 464 & 1,593 \\ 
	\hline 
  \end{tabular}
\end{center}
%\caption{Statistics of hand-based person Re-Id datasets used in this paper: 11k~\cite{Mah19} and HD~\cite{KumXu16}. Number of identities (ids), number of images and number of cameras are shown for train set, query set and gallery set of each dataset.}
%\label{tbl:HandDatasets}
\end{table*}
\noindent

\subsection{Implementation Details}

We implemented the LAGA-Net using PyTorch deep learning framework and trained it on NVIDIA GeForce RTX 2080 Ti GPU. 

\textbf{Body-based person Re-Id:} The input images of size $384 \times 128$ are resized to $\frac{9}{8}$ times of the size of the input images and then randomly cropped to  $384 \times 128$, augmented by random horizontal flip, normalization, color jittering and random erasing~\cite{ZhuLiaGuo20} during training. The test images are resized to $384 \times 128$ and augmented only by normalization.

\textbf{Hand-based person Re-Id:} The input images are resized to $356 \times 356$ and then randomly cropped to $324 \times 324$, augmented by random horizontal flip, normalization and color jittering during training. However, only normalization is utilized during testing with the test images resized to $324 \times 324$, without a random crop.

In both cases, a random order of images are used by reshuffling the dataset. We use a combination of cross-entropy and hard mining triplet losses over the 6 ID predictions as in Eq.~(\ref{eqn:totalLoss}) to train the LAGA-Net. To prevent over-fitting and over-confidence, label smoothing~\cite{SzeVanIof16} with smoothing value ($\epsilon$) of 0.1 is also used with the cross-entropy loss. We train the model for 70 epochs with mini-batch size of 20 and Adam optimizer with the weight decay factor for L2 regularization of $5 \times 10^{-4}$. For the first 10 epochs, we use a warmup strategy~\cite{LuoGuLia20}, increasing a learning rate linearly from $8 \times 10^{-6}$ to $8 \times 10^{-4}$, and then it is decayed to $4 \times 10^{-4}$, $2 \times 10^{-4}$ after 40 and 60 epochs, respectively. The learning rate is divided by 10 for the existing layers of the backbone network i.e. ten times bigger learning rate is given to the newly added layers (FC layers and batch normalizations) and the attention modules (embedding functions and batch normalizations), with appropriate weight and bias initializations.

\subsection{Evaluation Metrics}

We use the standard person Re-ID evaluation metrics, particularly, Cumulative Matching Characteristics (CMC)~\cite{ZheSheTia15} (rank-1 or top-1 matching accuracy) and mean Average Precision (mAP)~\cite{ZheSheTia15} to evaluate our proposed person Re-Id method.

\subsection{Ablation Analysis}  \label{sec:Ablation}

Our proposed method, as described in Section~\ref{sec:proposedMethod}, incorporates two complementary attention modules, channel and spatial with relative position, at a higher level of the network in branches along with the global (without attention) and local branches. The ablation analysis of these components is given in Table~\ref{tbl:Ablation} with evaluations on the Market-1501~\cite{ZheSheTia15} body dataset and palmar right (P-r) of the 11k hands dataset~\cite{Mah19}. As shown in this table, each component contributes to a performance gain. For body-based person Re-Id, for instance, the global branch (ResNet50 with some modifications) gives rank-1 and mAP of 93.92\% and 83.21\%, respectively. Incorporating the local branch with three horizontal partitions (stripes) boosts the performance to 94.41\% rank-1 and 85.30\% mAP. Rank-1 accuracy and mAP of 95.12\% and 86.91\%, respectively, are obtained by integrating the channel attention module (CAM). Incorporating the spatial attention module with relative positional encodings (SAM-RPE) contributes to the performance gain as well, giving the overall LAGA-Net performance of 96.18\% rank-1 and 88.76\% mAP on the Market-1501 dataset. Similarly, each component contributes to a performance gain on palmar right (P-r) of the 11k hands dataset as shown in this table.

%\begin{table}[htbp]%[!htb]%[!h]%[tb]
%\begin{center}
%\begin{tabular}{|c|c|c|}
%\hline
%    & rank-1 (\%) & mAP (\%) \\ 
%\hline
%Global + Local & 95.43 & 95.95  \\
% (ResNet50) &  &  \\ 
%%\hline 
%+ Spatial attention & 96.50 & 96.73 \\ 
%%\hline 
%+ Channel attention & 97.41 & 97.65 \\
%%\hline
%+ Relative position & \textbf{98.05} & \textbf{98.42} \\
%\hline 
%\end{tabular} 
%\end{center}
%\caption{Ablation analysis on components of MBA-Net on right palmar (P-r) of 11k hands dataset.}
%\label{tbl:Ablation}
%\end{table}
%\noindent

%\begin{table}[htbp]%[!htb]%[!h]%[tb]
%\begin{center}
%\begin{tabular}{|c|c|c|}
%\hline
%    & rank-1 (\%) & mAP (\%) \\ 
%\hline
%Global & 93.92 & 83.21 \\
%%\hline 
%+ Local & 94.41 & 85.30 \\ 
%%\hline 
%+ CAM & 95.12 & 86.91 \\
%%\hline
%+ SAM-RPE & \textbf{96.18} & \textbf{88.76} \\
%\hline 
%\end{tabular} 
%\end{center}
%\caption{Ablation analysis on components of LAGA-Net on Market-1501~\cite{ZheSheTia15}. Global + Local + CAM + SAM-RPE gives LAGA-Net.}
%\label{tbl:Ablation}
%\end{table}
%\noindent

\begin{table}[htbp]%[!htb]%[!h]%[tb]
\caption{\normalfont{Ablation analysis on components of LAGA-Net on Market-1501~\cite{ZheSheTia15} and palmar right (P-r) of 11k~\cite{Mah19}. Global + Local + CAM + SAM-RPE gives LAGA-Net. The results are shown in rank-1 accuracy (\%) and mAP (\%).}}
\label{tbl:Ablation}
\begin{center}
\begin{tabular}{|l|ll|ll|}
\hline
    \multirow{2}{*}{Method} & 
      \multicolumn{2}{c|}{Marke1501} & 
      \multicolumn{2}{c|}{P-r (11k)} \\
      \cline{2-5}
    & rank-1 & mAP & rank-1 & mAP\\ 
\hline
Global & 93.92 & 83.21 & 95.43 & 95.95 \\
%\hline 
+ Local & 94.41 & 85.30 & 96.52 & 96.97 \\ 
%\hline 
+ CAM & 95.12 & 86.91 & 97.14 & 97.45 \\
%\hline
+ SAM-RPE & \textbf{96.18} & \textbf{88.76} & \textbf{98.16} & \textbf{98.54} \\
\hline 
\end{tabular} 
\end{center}
%\caption{Ablation analysis on components of LAGA-Net on Market-1501~\cite{ZheSheTia15} and palmar right (P-r) of 11k~\cite{Mah19}. Global + Local + CAM + SAM-RPE gives LAGA-Net. The results are shown in rank-1 accuracy (\%) and mAP (\%).}
%\label{tbl:Ablation}
\end{table}
\noindent

%\subsection{Performance Evaluation}  
\subsection{Comparison with the State-of-the-art Methods}

\textbf{Body-based person Re-Id:} We evaluate our model and report the results using rank-1 matching accuracy and mAP~\cite{ZheSheTia15} on Market-1501~\cite{ZheSheTia15}, DukeMTMC-Re-ID~\cite{RisSolZou16}, CUHK03~\cite{WeiZhaXia14} and MSMT17~\cite{WeiZhaGao18} datasets. For fair comparison, we did not use post-processing such as re-ranking~\cite{ZhuLiaDon17} or multi-query~\cite{ZheSheTia15}. We compare our proposed method, LAGA-Net, to many existing state-of-the-art methods and report the quantitative performance comparison in Table~\ref{tbl:ComparisonBody}. As shown in this table, our method outperforms all other methods across all datasets in both rank-1 accuracy and mAP evaluation metrics except on CUHK03 where our method is ranked 2nd in rank-1 accuracy. This indicates that our method is more generalizable than the other methods across all datasets. Specifically, our proposed method outperforms the existing part-based methods such as PCB+RPP~\cite{YifLiaYi18} and MGN~\cite{WanYuaChe18} by large margin. For instance, the LAGA-Net outperforms the MGN by 13.73\% in rank-1 accuracy and 10.54\% mAP on CUHK03 dataset. Similarly, the LAGA-Net outperforms the existing attention-based methods such as MHN~\cite{BinWeiJia19}, ABD-Net~\cite{TiaShaJin19} and RGA-Net~\cite{ZhiCuiWen20}. For instance, the LAGA-Net outperforms the RGA-Net by 2.26\% in rank-1 accuracy and 3.47\% in mAP on MSMT17 dataset. Our proposed method outperforms not only these supervised body-based person Re-Id methods, but also the recent unsupervised person Re-Id methods such as RLCC~\cite{XiaYixYu21} and IICS~\cite{ShiShi21} as shown in Table~\ref{tbl:ComparisonBody}.

\begin{table*} [htbp]%[!htb]%[!h]%[tb]
\caption{\normalfont{Quantitative performance comparison of our method (LAGA-Net) with existing state-of-the-art body-based person Re-Id methods on Market-1501~\cite{ZheSheTia15}, DukeMTMC-Re-ID~\cite{RisSolZou16}, CUHK03~\cite{WeiZhaXia14} and MSMT17~\cite{WeiZhaGao18} datasets. The results are shown in rank-1  accuracy (\%) and mAP (\%). Best and second best results are shown in $\color{red}{\textbf{red}}$ and $\color{blue}{\textbf{blue}}$, respectively. * denotes unsupervised person Re-Id methods.}}
\label{tbl:ComparisonBody}
\begin{center}
  \begin{tabular}{|l|ll|ll|ll|ll|}
    \hline
    \multirow{2}{*}{Method} & 
      \multicolumn{2}{c|}{Market-1501} & 
      \multicolumn{2}{c|}{DukeMTMC}    &
      \multicolumn{2}{c|}{CUHK03 (L)}  &
      \multicolumn{2}{c|}{MSMT17}     \\
      \cline{2-9}
    & rank-1 & mAP & rank-1 & mAP  & rank-1 & mAP & rank-1 & mAP \\
    \hline
    MGCAM~\cite{SonHuaOuy18} & 83.79 & 74.33 & - & - & 50.14 & 50.21 & - & - \\
    DGNet~\cite{ZheXiaZhi19} & 94.8 & 86.0 & 86.6 & 74.8 & - & - & 77.2 & 52.3 \\
    Interpreter-50~\cite{XiaXinWu21} & 94.74 & 87.11 & 87.84 & 75.27 & - & - & - & - \\
    OSNet~\cite{KaiYonAnd19} & 94.8 & 84.9 & 88.6 & 73.5 & 72.3 & 67.8  & 78.7 & 52.9 \\ 
    MGN~\cite{WanYuaChe18} & 95.7 & 86.9 & 88.7 & 78.4 & 68.0 & 67.4 & - & - \\
    PCB+RPP~\cite{YifLiaYi18} & 93.8 & 81.6 & 83.3 & 69.2 & - & - & 68.2 & 40.4 \\
    PDC~\cite{SuLiZha17}  & 84.14 & 63.41 & - & - & 88.70 & - & - & - \\ 
    BagTicks~\cite{LuoGuLia20} & 94.5 & 85.9 & 86.4 & 76.4 & - & - & - & -  \\
	ABD-Net~\cite{TiaShaJin19} & 95.60 & 88.28 & 89.00 & \color{blue}{\textbf{78.59}}  & - & - & \color{blue}{\textbf{82.30}} & \color{blue}{\textbf{60.80}} \\
	RGA-Net~\cite{ZhiCuiWen20} & \color{blue}{\textbf{96.10}} & \color{blue}{\textbf{88.40}} & - & - & 81.10 & 77.40 & 80.30 & 57.50 \\ 
	MHN~\cite{BinWeiJia19} & 95.1 & 85.0 & \color{blue}{\textbf{89.1}} & 77.2 & 77.2 & 72.4 & - & - \\
	IANet~\cite{RuiBinHon19} & 94.4 & 83.1 & 87.1 & 73.4 & \color{red}{\textbf{92.4}} & - & 75.5 &  46.8 \\
	RLCC*~\cite{XiaYixYu21} & 90.8 & 77.7 & 83.2 & 69.2 & - & - & 56.5 & 27.9 \\
	IICS*~\cite{ShiShi21} & 88.8 & 72.1 & 80.0 & 64.4 & - & - & 56.4 & 26.9 \\
	%\hline
	\textbf{LAGA-Net (Ours)} & \color{red}{\textbf{96.18}} & \color{red}{\textbf{88.76}} & \color{red}{\textbf{89.71}} & \color{red}{\textbf{78.92}} & \color{blue}{\textbf{81.73}} & \color{red}{\textbf{77.94}} & \color{red}{\textbf{82.56}} & \color{red}{\textbf{60.97}} \\
	\hline 
  \end{tabular}
\end{center}
%\caption{Quantitative performance comparison of our method (LAGA-Net) with other state-of-the-art methods on Market-1501~\cite{ZheSheTia15}, DukeMTMC-Re-ID~\cite{RisSolZou16}, CUHK03~\cite{WeiZhaXia14} and MSMT17~\cite{WeiZhaGao18} datasets. The results are shown in rank-1  accuracy (\%) and mAP (\%). Best and second best results are shown in $\color{red}{\textbf{red}}$ and $\color{blue}{\textbf{blue}}$, respectively. * denotes unsupervised person Re-Id methods.}
%\label{tbl:ComparisonBody}
\end{table*}
\noindent

\textbf{Hand-based person Re-Id:} We compare our proposed method to many existing state-of-the-art hand-based Re-Id (recognition) methods such as GPA-Net~\cite{BaiWilHosGPA21}, MBA-Net~\cite{BaiWilHosMBA21}, RGA-Net~\cite{ZhiCuiWen20} and ABD-Net~\cite{TiaShaJin19}. The GPA-Net was designed for hand-based person identification, however, both RGA-Net and ABD-Net were designed for body-based person re-identification. Therefore, we trained both RGA-Net and ABD-Net on hand datasets using the same experimental settings (loss function, optimizer, hyperparameters, etc.) as our method to make a fair comparasion with our method. The quantitative performance comparison of our method with the other methods is given in Table~\ref{tbl:ComparisonHand}. As shown in this table, our method outperforms all other methods across all datasets in both rank-1 accuracy and mAP evaluation metrics. For instance, the LAGA-Net outperforms the MBA-Net by 0.81\% in rank-1 accuracy and 0.73\% in mAP on HD dataset. The adapted body-based person Re-Id methods, RGA-Net~\cite{ZhiCuiWen20} and ABD-Net~\cite{TiaShaJin19}, have limited performance on hand datasets. For instance, our method outperfoms the RGA-Net by 5.50\% in rank-1 accuracy and 4.96\% in mAP on palmar right (P-r) of 11k dataset.

\begin{table*} [htbp]%[!htb]%[!h]%[tb]
\caption{\normalfont{Quantitative performance comparison of our method (LAGA-Net) with existing state-of-the-art hand-based person Re-Id methods (GPA-Net~\cite{BaiWilHosGPA21}, MBA-Net~\cite{BaiWilHosMBA21}, RGA-Net~\cite{ZhiCuiWen20} and ABD-Net~\cite{TiaShaJin19}) on right dorsal (D-r) of 11k, left dorsal (D-l) of 11k, right palmar (P-r) of 11k, left palmar (P-l) of 11k and HD datasets. The results are shown in rank-1  accuracy (\%) and mAP (\%). Best and second best results are shown in $\color{red}{\textbf{red}}$ and $\color{blue}{\textbf{blue}}$, respectively.}}
\label{tbl:ComparisonHand}
\begin{center}
  \begin{tabular}{|l|ll|ll|ll|ll|ll|}
    \hline
    \multirow{2}{*}{Method} & 
      \multicolumn{2}{c|}{D-r of 11k} & 
      \multicolumn{2}{c|}{D-l of 11k} & 
      \multicolumn{2}{c|}{P-r of 11k} &
      \multicolumn{2}{c|}{P-l of 11k} &
      \multicolumn{2}{c|}{HD} \\
      \cline{2-11}
    & rank-1 & mAP & rank-1 & mAP & rank-1 & mAP & rank-1 & mAP & rank-1 & mAP \\
    \hline
	GPA-Net~\cite{BaiWilHosGPA21} & 94.80 & 95.72 & 94.87 & 95.93 & 95.83 & 96.31 & 95.72 & 96.20 & 94.64 & 95.08 \\ 
	MBA-Net~\cite{BaiWilHosMBA21} & \color{blue}{\textbf{97.45}} & \color{blue}{\textbf{97.98}} & \color{blue}{\textbf{96.71}} & \color{blue}{\textbf{97.41}} & \color{blue}{\textbf{98.05}} & \color{blue}{\textbf{98.42}} & \color{blue}{\textbf{97.42}} & \color{blue}{\textbf{97.84}} & \color{blue}{\textbf{95.12}} & \color{blue}{\textbf{95.54}} \\
	%\hline 
	RGA-Net~\cite{ZhiCuiWen20} & 94.77 & 95.67 & 95.30 & 95.98 & 92.66 & 93.58 & 94.95 & 95.67 & 95.06 & 95.39 \\ 
	%\hline 
	ABD-Net~\cite{TiaShaJin19} & 95.89 & 96.76 & 94.26 & 95.34 & 96.21 & 96.91 & 95.54 & 96.01 & 94.93 & 95.38 \\
	%\hline
	\textbf{LAGA-Net (Ours)} & \color{red}{\textbf{97.56}} & \color{red}{\textbf{98.11}} & \color{red}{\textbf{96.82}} & \color{red}{\textbf{97.53}} & \color{red}{\textbf{98.16}} & \color{red}{\textbf{98.54}} & \color{red}{\textbf{97.56}} & \color{red}{\textbf{97.95}} & \color{red}{\textbf{95.93}} & \color{red}{\textbf{96.27}} \\
	\hline 
  \end{tabular}
\end{center}
%\caption{Quantitative performance comparison of our method (LAGA-Net) with other methods (GPA-Net~\cite{BaiWilHosGPA21}, MBA-Net~\cite{BaiWilHosMBA21}, RGA-Net~\cite{ZhiCuiWen20} and ABD-Net~\cite{TiaShaJin19}) on right dorsal (D-r) of 11k, left dorsal (D-l) of 11k, right palmar (P-r) of 11k, left palmar (P-l) of 11k and HD datasets. The results are shown in rank-1  accuracy (\%) and mAP (\%). Best and second best results are shown in $\color{red}{\textbf{red}}$ and $\color{blue}{\textbf{blue}}$, respectively.}
%\label{tbl:ComparisonHand}
\end{table*}
\noindent

\subsection{Qualitative Re-ID Results}

\begin{figure}[htbp]%[!htb] %[t]%[!h]
  \begin{center}
  %\subfloat[] %[Channel Attention Module (CAM)]
  {\label{fig:p1} \includegraphics[width=0.80\linewidth]{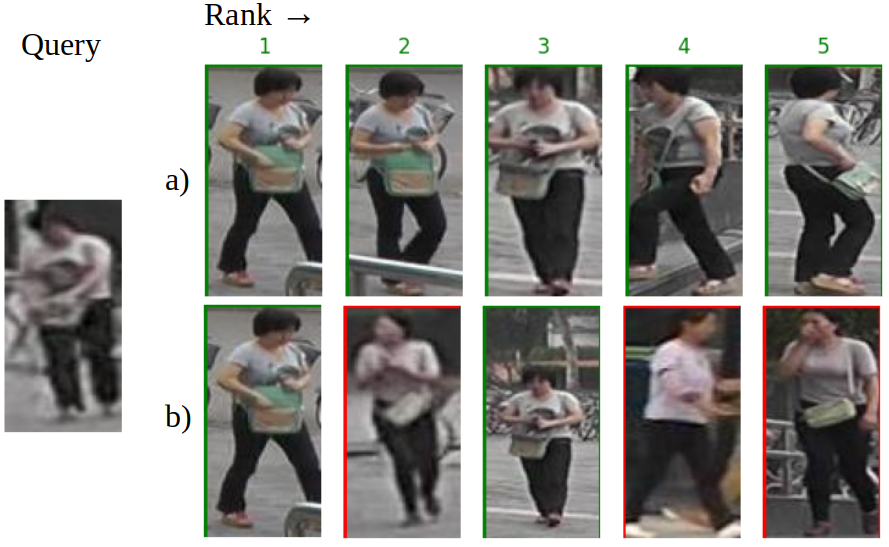}} \\%& %\\ %height=0.46
  %\subfloat[] %[Spatial Attention Module with Relative Positional Encodings (SAM-RPE)]
  {\label{fig:p2} \includegraphics[width=0.80\linewidth]{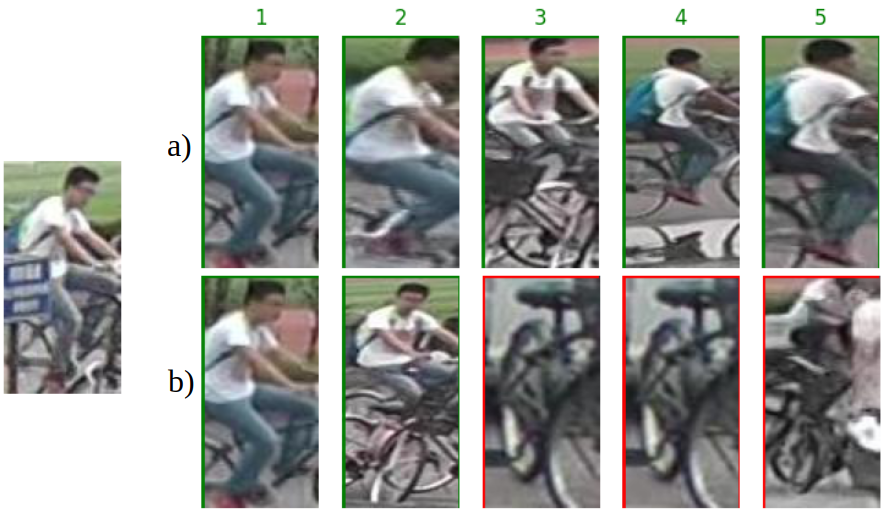}}  \\ %height=0.18
    %\subfloat[] %[Spatial Attention Module with Relative Positional Encodings (SAM-RPE)]
  {\label{fig:p3} \includegraphics[width=0.80\linewidth]{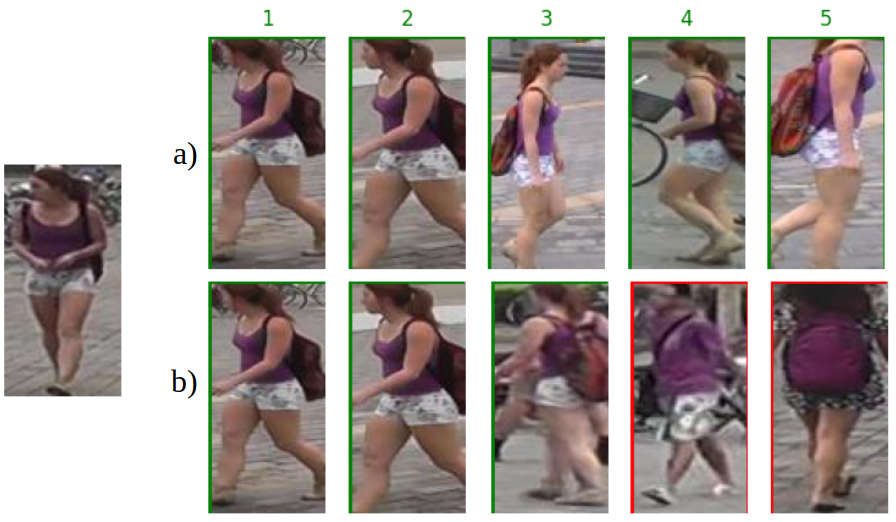}}  %& %\\ %height=0.18
  \end{center}
   \caption{Some qualitative results of our method on Market-1501~\cite{ZheSheTia15} dataset using query vs ranked results retrieved from gallery. Left: query image, Right: a) top-5 results of the LAGA-Net, b) top-5 results of the global (without attention) component of the LAGA-Net (baseline). The green and red bounding boxes denote the correct and the wrong matches, respectively. Feature embeddings from our proposed method (LAGA-Net) give better retrieval performance.}
  \label{fig:demoResultP}
\end{figure}

\begin{figure}[htbp] %[t]%[!htb] %[t]%[!h]
\begin{center}
  \includegraphics[width=1.0\linewidth]{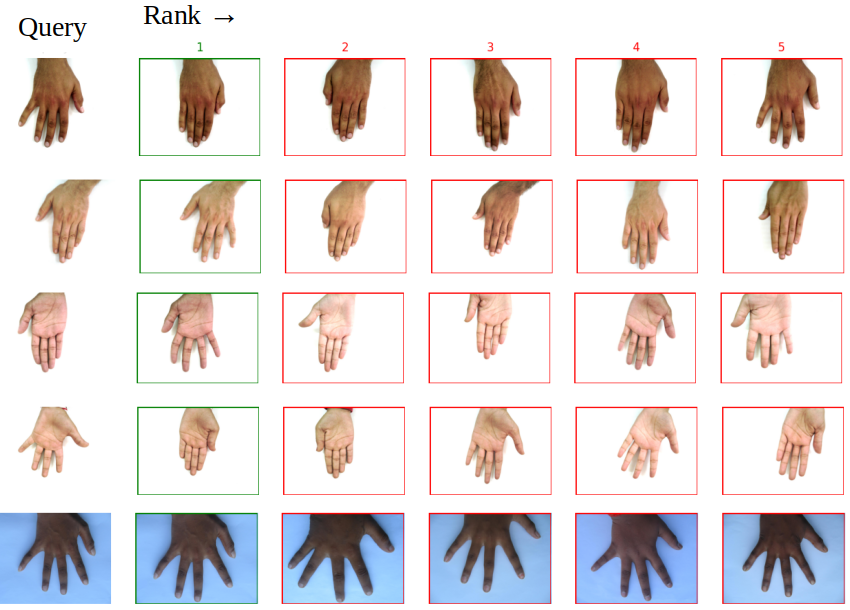} \\%& %\\ %height=0.46
\end{center}
   \caption{Some qualitative results of our method on 11k~\cite{Mah19} and HD~\cite{KumXu16} datasets using query vs ranked results retrieved from gallery. From top to bottom row are right dorsal of 11k, left dorsal of 11k, right palmar of 11k, left palmar of 11k and HD datasets. The green and red bounding boxes denote the correct and the wrong matches, respectively.}
\label{fig:demoResultH}
\end{figure}

The qualitative Re-ID results of our proposed method is shown in Figs.~\ref{fig:demoResultP} and~\ref{fig:demoResultH}. As can be observed on Fig.~\ref{fig:demoResultP}, our proposed method (LAGA-Net) has an improved performance on Market-1501 dataset over the baseline (the global without attention component of the LAGA-Net) in retrieval performance. While the LAGA-Net retrieves all top-5 correct results (top row for each query), the baseline only retrieves few results (bottom row for each query). This indicates that the LAGA-Net learns more robust discriminative feature embeddings which help to find more correct (true positive) results than the baseline even when the persons in the images are under significant appearance variations. We also show a qualitative exemplar image (query) of each (sub-)dataset of the hand datasets in Fig.~\ref{fig:demoResultH} with ranked results retrieved from a gallery of each (sub-)dataset. There is only one correct image in the galleries of the hand-based person Re-Id datasets unlike in the body-based person Re-Id datasets. Bacause of this, only one correct image is retrieved from the gallery of the 11k and HD hand datasets for each query image whereas multiple correct images are retrieved from the Market-1501 gallery for each query image. Overall, our proposed method is effective on both the body-based and the hand-based person Re-Id datasets as it learns more robust and discriminative deep feature representations.

\noindent

\noindent

\section{Discussion}  \label{sec:Discussion}
The proposed multi-branch deep network architecture, LAGA-Net, is a compound approach for end-to-end discriminative deep feature representations learning for person Re-Id based on both body and hand images. We have shown in the experiments section (see Section~\ref{sec:experimentalResults}) that extensive evaluation of this method has demonstrated excellent performance not only on body datasets but also on hands datasets, as shown in Table~\ref{tbl:ComparisonBody} and Table~\ref{tbl:ComparisonHand}, respectively.

The different branches in the proposed multi-branch deep network capture different but complementary information to boost the performance of the network as shown in ablation analysis section (see Section~\ref{sec:Ablation}) on both body and hands datasets, particularly in Table~\ref{tbl:Ablation}. Each branch has its own importance in the proposed network. The attention branches, channel attention branch and spatial attention branch, focus on the relevant features of the image while suppressing the irrelevant backgrounds. To maintain translation equivariance, relative positional encodings is integrated into spatial attention module of the spatial attention branch. The global and local branches intends to capture global context and fine-grained information, respectively. By properly integrating these branches in our proposed network, we have demonstrated that it is possible to effectively learn robust and discriminative feature representations for person Re-Id based on both body and hand images. The main importance of the proposed method is the excellent performance it has shown on both body and hand images. The elegant approach of integrating many different components into our proposed compound approach has demonstrated learning of robust and discriminative feature representations which is helpful in overcoming the many challenges the person Re-Id is facing such as pose variations, occlusion, view point changes, lighting changes, background clutter, noisy labels, etc.

The proposed method has various applications. For instance, the person Re-ID based on body can be used for intelligent video surveillance in less controlled or uncontrolled environments. The hand-based person Re-Id can be used for criminal investigation in uncontrolled environments, for instance, for recognizing or re-identifying perpetrators of serious crime such as sexual abuse in case only hand images of the perpetrators are available, which is very crucial in assisting international police forces.

\section{Conclusion}  \label{sec:Conclusion}

In this work, we introduce a compound approach for end-to-end discriminative deep feature learning, the Local-Aware Global Attention Network (LAGA-Net), for person Re-Id based on both body and hand images. The LAGA-Net is a multi-branch deep network architecture consisting of channel and spatial attention modules in branches in addition to global (without attention) and local branches to learn deep attentive, global and part-level feature embeddings for more robust and discriminative person Re-Id. We also integrate relative positional encodings into the spatial attention module to capture the spatial positions of pixels to overcome the weakness of the attention mechanisms, equivariant to pixel shuffling. The incorporation of these branches allows a deeper study of the features of the person body and hand images for robust re-identification of individuals in less controlled and challenging environments. The LAGA-Net demonstrates the state-of-the-art performance through extensive experiments on four popular body-based person Re-Id benchmarks and two publicly available hand datasets where the ablation analysis shows each component substantially contributes to a performance gain.

%\section*{Acknowledgment}
%
%We would like to thank Bryan Williams, Hossein Rahmani, Plamen Angelov and Sue Black who were part of the project which our preliminary work~\cite{BaiWilHosMBA21} is associated to.

% References should be produced using the bibtex program from suitable
% BiBTeX files (here: strings, refs, manuals). The IEEEbib.bst bibliography
% style file from IEEE produces unsorted bibliography list.
% -------------------------------------------------------------------------

\bibliographystyle{IEEEbib}
\bibliography{refs}

\begin{thebibliography}{10}

\bibitem{Nat21}
Nathanael~L. Baisa,
\newblock ``Occlusion-robust online multi-object visual tracking using a
  {GM-PHD} filter with {CNN}-based re-identification,''
\newblock {\em Journal of Visual Communication and Image Representation}, vol.
  80, pp. 103279, 2021.

\bibitem{RisTom18}
Ergys Ristani and Carlo Tomasi,
\newblock ``Features for multi-target multi-camera tracking and
  re-identification,''
\newblock in {\em {IEEE} Conference on Computer Vision and Pattern Recognition
  ({CVPR}), Salt Lake City, UT, USA, June 18-22}, 2018, pp. 6036--6046.

\bibitem{WeiZhaGao18}
Longhui Wei, Shiliang Zhang, Wen Gao, and Qi~Tian,
\newblock ``Person transfer {GAN} to bridge domain gap for person
  re-identification,''
\newblock in {\em Proceedings of the IEEE Conference on Computer Vision and
  Pattern Recognition (CVPR)}, June 2018.

\bibitem{YeSheLin21}
Mang Ye, Jianbing Shen, Gaojie Lin, Tao Xiang, Ling Shao, and Steven~C.H. Hoi,
\newblock ``Deep learning for person re-identification: A survey and outlook,''
\newblock {\em IEEE Transactions on Pattern Analysis and Machine Intelligence},
  pp. 1--1, 2021.

\bibitem{WeiZhaXia14}
Wei Li, Rui Zhao, Tong Xiao, and Xiaogang Wang,
\newblock ``Deepreid: Deep filter pairing neural network for person
  re-identification,''
\newblock in {\em 2014 IEEE Conference on Computer Vision and Pattern
  Recognition}, 2014, pp. 152--159.

\bibitem{ZheXiaZhi19}
Zhedong Zheng, Xiaodong Yang, Zhiding Yu, Liang Zheng, Yi~Yang, and Jan Kautz,
\newblock ``Joint discriminative and generative learning for person
  re-identification,''
\newblock in {\em 2019 IEEE/CVF Conference on Computer Vision and Pattern
  Recognition (CVPR)}, 2019, pp. 2133--2142.

\bibitem{LuoGuLia20}
Hao Luo, Wei Jiang, Youzhi Gu, Fuxu Liu, Xingyu Liao, Shenqi Lai, and Jianyang
  Gu,
\newblock ``A strong baseline and batch normalization neck for deep person
  re-identification,''
\newblock {\em IEEE Transactions on Multimedia}, vol. 22, no. 10, pp.
  2597--2609, 2020.

\bibitem{SuLiZha17}
C.~{Su}, J.~{Li}, S.~{Zhang}, J.~{Xing}, W.~{Gao}, and Q.~{Tian},
\newblock ``Pose-driven deep convolutional model for person
  re-identification,''
\newblock in {\em 2017 IEEE International Conference on Computer Vision
  (ICCV)}, 2017, pp. 3980--3989.

\bibitem{YifLiaYi18}
Yifan Sun, Liang Zheng, Yi~Yang, Qi~Tian, and Shengjin Wang,
\newblock ``Beyond part models: Person retrieval with refined part pooling (and
  a strong convolutional baseline),''
\newblock in {\em Proceedings of the European Conference on Computer Vision
  (ECCV)}, September 2018.

\bibitem{WanYuaChe18}
Guanshuo Wang, Yufeng Yuan, Xiong Chen, Jiwei Li, and Xi~Zhou,
\newblock ``Learning discriminative features with multiple granularities for
  person re-identification,''
\newblock in {\em Proceedings of the 26th ACM International Conference on
  Multimedia}, New York, NY, USA, 2018, MM '18, p. 274–282, Association for
  Computing Machinery.

\bibitem{BinWeiJia19}
Binghui Chen, Weihong Deng, and Jiani Hu,
\newblock ``Mixed high-order attention network for person re-identification,''
\newblock in {\em 2019 IEEE/CVF International Conference on Computer Vision
  (ICCV)}, 2019, pp. 371--381.

\bibitem{TiaShaJin19}
Tianlong Chen, Shaojin Ding, Jingyi Xie, Ye~Yuan, Wuyang Chen, Yang Yang, Zhou
  Ren, and Zhangyang Wang,
\newblock ``{ABD-Net}: Attentive but diverse person re-identification,''
\newblock in {\em 2019 IEEE/CVF International Conference on Computer Vision
  (ICCV)}, 2019, pp. 8350--8360.

\bibitem{ZhiCuiWen20}
Zhizheng Zhang, Cuiling Lan, Wenjun Zeng, Xin Jin, and Zhibo Chen,
\newblock ``Relation-aware global attention for person re-identification,''
\newblock in {\em 2020 IEEE/CVF Conference on Computer Vision and Pattern
  Recognition (CVPR)}, 2020, pp. 3183--3192.

\bibitem{JaiDebEng22}
Anil~K. Jain, Debayan Deb, and Joshua~J. Engelsma,
\newblock ``Biometrics: Trust, but verify,''
\newblock {\em IEEE Transactions on Biometrics, Behavior, and Identity
  Science}, vol. 4, no. 3, pp. 303--323, 2022.

\bibitem{DanEliRos16}
A.~{Dantcheva}, P.~{Elia}, and A.~{Ross},
\newblock ``What else does your biometric data reveal? {A} survey on soft
  biometrics,''
\newblock {\em IEEE Transactions on Information Forensics and Security}, vol.
  11, no. 3, pp. 441--467, 2016.

\bibitem{BaiWilHosGPA21}
Nathanael~L. Baisa, Bryan Williams, Hossein Rahmani, Plamen Angelov, and Sue
  Black,
\newblock ``Hand-based person identification using global and part-aware deep
  feature representation learning,''
\newblock in {\em 2022 Eleventh International Conference on Image Processing
  Theory, Tools and Applications (IPTA)}, 2022, pp. 1--6.

\bibitem{YimChaShu20}
Yimin Yuan, Chaoying Tang, Shuhang Xia, Zhou Chen, and Tong Qi,
\newblock ``{HandNet}: Identification based on hand images using deep learning
  methods,''
\newblock in {\em Proceedings of the 2020 4th International Conference on
  Vision, Image and Signal Processing}, New York, NY, USA, 2020, ICVISP 2020,
  Association for Computing Machinery.

\bibitem{AttAkhCha21}
Abdelouahab Attia, Zahid Akhtar, and Youssef Chahir,
\newblock ``{Feature-level fusion of major and minor dorsal finger knuckle
  patterns for person authentication},''
\newblock {\em {Signal, Image and Video Processing}}, Feb. 2021.

\bibitem{ZheSheTia15}
L.~{Zheng}, L.~{Shen}, L.~{Tian}, S.~{Wang}, J.~{Wang}, and Q.~{Tian},
\newblock ``Scalable person re-identification: A benchmark,''
\newblock in {\em 2015 IEEE International Conference on Computer Vision
  (ICCV)}, 2015, pp. 1116--1124.

\bibitem{RisSolZou16}
Ergys Ristani, Francesco Solera, Roger Zou, Rita Cucchiara, and Carlo Tomasi,
\newblock ``Performance measures and a data set for multi-target, multi-camera
  tracking,''
\newblock in {\em Computer Vision -- ECCV 2016 Workshops}, Gang Hua and
  Herv{\'e} J{\'e}gou, Eds., Cham, 2016, pp. 17--35, Springer International
  Publishing.

\bibitem{Mah19}
Mahmoud Afifi,
\newblock ``11k hands: gender recognition and biometric identification using a
  large dataset of hand images,''
\newblock {\em Multimedia Tools and Applications}, 2019.

\bibitem{KumXu16}
A.~{Kumar} and Z.~{Xu},
\newblock ``Personal identification using minor knuckle patterns from palm
  dorsal surface,''
\newblock {\em IEEE Transactions on Information Forensics and Security}, vol.
  11, no. 10, pp. 2338--2348, 2016.

\bibitem{XiaYixYu21}
Xiao Zhang, Yixiao Ge, Yu~Qiao, and Hongsheng Li,
\newblock ``Refining pseudo labels with clustering consensus over generations
  for unsupervised object re-identification,''
\newblock in {\em 2021 IEEE/CVF Conference on Computer Vision and Pattern
  Recognition (CVPR)}, 2021, pp. 3435--3444.

\bibitem{ShiShi21}
Shiyu Xuan and Shiliang Zhang,
\newblock ``Intra-inter camera similarity for unsupervised person
  re-identification,''
\newblock in {\em 2021 IEEE/CVF Conference on Computer Vision and Pattern
  Recognition (CVPR)}, 2021, pp. 11921--11930.

\bibitem{SonHuaOuy18}
Chunfeng Song, Yan Huang, Wanli Ouyang, and Liang Wang,
\newblock ``Mask-guided contrastive attention model for person
  re-identification,''
\newblock in {\em 2018 IEEE/CVF Conference on Computer Vision and Pattern
  Recognition}, 2018, pp. 1179--1188.

\bibitem{KhaNasHay21}
Salman Khan, Muzammal Naseer, Munawar Hayat, Syed~Waqas Zamir, Fahad~Shahbaz
  Khan, and Mubarak Shah,
\newblock ``Transformers in vision: A survey,'' 2021.

\bibitem{ShaJakAsh18}
Peter Shaw, Jakob Uszkoreit, and Ashish Vaswani,
\newblock ``Self-attention with relative position representations,''
\newblock in {\em Proceedings of the 2018 Conference of the North {A}merican
  Chapter of the Association for Computational Linguistics: Human Language
  Technologies, Volume 2 (Short Papers)}, New Orleans, Louisiana, June 2018,
  pp. 464--468, Association for Computational Linguistics.

\bibitem{SheBelVem20}
Zhuoran Shen, Irwan Bello, Raviteja Vemulapalli, Xuhui Jia, and Ching{-}Hui
  Chen,
\newblock ``Global self-attention networks for image recognition,''
\newblock {\em CoRR}, vol. abs/2010.03019, 2020.

\bibitem{HeZhaSun16}
K.~{He}, X.~{Zhang}, S.~{Ren}, and J.~{Sun},
\newblock ``Deep residual learning for image recognition,''
\newblock in {\em 2016 IEEE Conference on Computer Vision and Pattern
  Recognition (CVPR)}, 2016, pp. 770--778.

\bibitem{ChrSerVin17}
Christian Szegedy, Sergey Ioffe, Vincent Vanhoucke, and Alexander~A. Alemi,
\newblock ``Inception-v4, inception-resnet and the impact of residual
  connections on learning,''
\newblock in {\em Proceedings of the Thirty-First AAAI Conference on Artificial
  Intelligence}. 2017, AAAI'17, p. 4278–4284, AAAI Press.

\bibitem{HuaLiuVan17}
G.~Huang, Z.~Liu, L.~Van~Der Maaten, and K.~Q. Weinberger,
\newblock ``Densely connected convolutional networks,''
\newblock in {\em 2017 IEEE Conference on Computer Vision and Pattern
  Recognition (CVPR)}, Los Alamitos, CA, USA, jul 2017, pp. 2261--2269, IEEE
  Computer Society.

\bibitem{SzeVanIof16}
C.~{Szegedy}, V.~{Vanhoucke}, S.~{Ioffe}, J.~{Shlens}, and Z.~{Wojna},
\newblock ``Rethinking the inception architecture for computer vision,''
\newblock in {\em 2016 IEEE Conference on Computer Vision and Pattern
  Recognition (CVPR)}, 2016, pp. 2818--2826.

\bibitem{AlexLucBas17}
Alexander Hermans, Lucas Beyer, and Bastian Leibe,
\newblock ``In defense of the triplet loss for person re-identification,''
  2017.

\bibitem{ZhuLiaDon17}
Zhun Zhong, Liang Zheng, Donglin Cao, and Shaozi Li,
\newblock ``Re-ranking person re-identification with k-reciprocal encoding,''
\newblock in {\em 2017 IEEE Conference on Computer Vision and Pattern
  Recognition (CVPR)}, 2017, pp. 3652--3661.

\bibitem{ZhuLiaGuo20}
Zhun Zhong, Liang Zheng, Guoliang Kang, Shaozi Li, and Yi~Yang,
\newblock ``Random erasing data augmentation,''
\newblock in {\em The Thirty-Fourth {AAAI} Conference on Artificial
  Intelligence, {AAAI} 2020, The Thirty-Second Innovative Applications of
  Artificial Intelligence Conference, {IAAI} 2020, The Tenth {AAAI} Symposium
  on Educational Advances in Artificial Intelligence, {EAAI} 2020, New York,
  NY, USA, February 7-12}. 2020, pp. 13001--13008, {AAAI} Press.

\bibitem{XiaXinWu21}
Xiaodong Chen, Xinchen Liu, Wu~Liu, Xiao-Ping Zhang, Yongdong Zhang, and Tao
  Mei,
\newblock ``Explainable person re-identification with attribute-guided metric
  distillation,''
\newblock in {\em Proceedings of the IEEE/CVF International Conference on
  Computer Vision (ICCV)}, October 2021, pp. 11813--11822.

\bibitem{KaiYonAnd19}
Kaiyang Zhou, Yongxin Yang, Andrea Cavallaro, and Tao Xiang,
\newblock ``Omni-scale feature learning for person re-identification,''
\newblock in {\em 2019 IEEE/CVF International Conference on Computer Vision
  (ICCV)}, 2019, pp. 3701--3711.

\bibitem{RuiBinHon19}
Ruibing Hou, Bingpeng Ma, Hong Chang, Xinqian Gu, Shiguang Shan, and Xilin
  Chen,
\newblock ``Interaction-and-aggregation network for person re-identification,''
\newblock in {\em 2019 IEEE/CVF Conference on Computer Vision and Pattern
  Recognition (CVPR)}, 2019, pp. 9309--9318.

\bibitem{BaiWilHosMBA21}
Nathanael~L. Baisa, Bryan Williams, Hossein Rahmani, Plamen Angelov, and Sue
  Black,
\newblock ``Multi-branch with attention network for hand-based person
  recognition,''
\newblock in {\em 2022 26th International Conference on Pattern Recognition
  (ICPR)}, 2022.

\end{thebibliography}

\vfill

\end{document}